\pdfoutput=1
\documentclass[11pt,a4paper]{article}

\makeatletter                   
\def\mdseries@tt{m}             
\makeatother                    

\usepackage{times}
\usepackage[hyperref]{acl2019}
\usepackage{latexsym}
\usepackage{kantlipsum}
\usepackage{amsmath}
\usepackage{graphicx}
\usepackage[utf8]{inputenc}

\usepackage{url}

\aclfinalcopy 

\title{Anaphora Resolution in Dialogue Systems for South Asian Languages}

\author{Vinay Annam \\
  BML Munjal University \\
  \texttt{annam.vinay.15csc} \\
  \texttt{@bml.edu.in}\\\And
  Nikhil Koditala\\
  BML Munjal University \\
  \texttt{nikhil.koditala.15cse} \\
  \texttt{@bml.edu.in}\\\And
  Radhika Mamidi\\
  IIIT Hyderabad\\
  \texttt{radhika.mamidi} \\
  \texttt{@iiit.ac.in}\\}

\date{}

\begin{document}
\maketitle
\begin{abstract}
Anaphora resolution is a challenging task which has been the interest of NLP researchers for a long time. Traditional resolution techniques like eliminative constraints and weighted preferences were successful in many languages. However, they are ineffective in free word order languages like most South Asian languages. Heuristic and rule-based techniques were typical in these languages, which are constrained to context and domain. In this paper, we venture a new strategy using neural networks for resolving anaphora in human-human dialogues. The architecture chiefly consists of three components, a shallow parser for extracting features, a feature vector generator which produces the word embeddings, and a neural network model which will predict the antecedent mention of an anaphora. The system has been trained and tested on Telugu conversation corpus we generated. Given the advantage of the semantic information in word embeddings and appending actor, gender, number, person and part of plural features the model has reached an F1-score of 86. 
\end{abstract}

\section{Introduction}

Throughout the information era, we have seen a shift in human-computer interactions, from clicks to chats. Conversational agents and dialogue systems are becoming prominent with the daily advances in the field of Artificial Intelligence. Technology will be effective if it can reach for the vaster population, by building computational models for popular languages. According to \cite{Ethnologue:19}, Telugu, which belongs to the Dravidian family, is one of the active growing languages and is ranked 16 among  7,111 living languages with 93 million speakers universally. Despite such attention, Telugu has inadequate resources when compared to its counter-partners. And also, with the advent of deep learning, many recent works are producing promising results for many languages.

In a discourse, anaphora is a lexical device which acts as a substitution for an entity mentioned earlier. As shown in example (\ref{ex1}) it is complicated to define a computable representation of the resolution process because humans personally deal with it subconsciously and mostly oblivious of the particularities. 
\begin{description}
\label{ex1}
\setlength{\itemsep}{0pt}
  \setlength{\parskip}{0pt}
  \setlength{\parsep}{0pt}
\item {
Example 1:

Shyam: Will Ram come to our school tomorrow for the competition?

Prem: It is too far from his house.

Here the pronouns 'his' refers to Ram, 'it' refers to school and ‘our’ refers to both Shyam and Prem.}
\end{description}

\noindent Despite the involvement of such intricacy, these systems are yet crucial in dialogue systems, machine translation, and information extraction. In this paper, we build a system that resolves the anaphora in Telugu dialogues. In contrast to syntactic and rule-based systems, which are approximate solutions, our method uses few handcrafted features appended to the word embeddings, focusing on semantic features and works excellently on real conversations. We present a new strategy to resolve speaker-hearer mentions and plural mentions, which were never tackled before. To the best of our knowledge, it is the first time deep learning has successfully implemented in Telugu dialogue NLP research.

\section{Related Work}
\citet{Hobbs:78} was one of the first persons to pioneer in the area of anaphora resolution focusing on early syntactic heuristics. His algorithm takes sentences up to target pronoun as input, and as it traverses backward it finds the noun phrases with same gender and number. Hobbs evaluated his algorithm manually and reported an accuracy of 88.3 percent. Then \cite{Hirst:81} directed the anaphora problem towards resolving it in discourse. \cite{leass:94,Denber:98} described several syntactic heuristics for reflexive, reciprocal and pleonastic anaphora. \cite{Grosz:95} claimed that at any given point there is a single entity being centered. Using this claim they proposed a centering algorithm which finds an entity which is divergent from other evoked entities. \cite{Mitkov:98} proposed a robust, knowledge-poor multilingual approach in resolving pronouns where each entity is provided a score based on indicators and entity with high score is considered antecedent. \cite{Ng:02} suggested a machine learning approach to anaphora resolution. However, statistical learning methods suffer from the difficulties of small corpora and corpus dependent learning. 

Most of the work in Indian languages has been done in Hindi, Bengali, and Tamil. \cite{Praveen:2013} built a hybrid approach for anaphora resolution in Hindi using dependency parser and a decision tree classifier. \cite{Hemanth:15} proposed a rule-based system for anaphora resolution in Telugu dialog systems, After preprocessing the data using Morphological analyzer and POS tagger they used a set of hard-coded rules to deal with different types of pronouns.

\citet{Clark:2015} has done pioneering work in coreference resolution using deep learning that automatically learns dense vector representations for mention pairs for English and Chinese. He built them using the word embeddings in the mention and surrounding context, which will maintain the semantic similarity. Despite using a few hand-engineered features, he trained an incremental coreference system that can utilize entity-level information. His mention pair model acted as an inspiration for our feature representations, and we updated it for free word order languages. In free word order languages, despite changing the order of words in a sentence the overall meaning of the sentence will not change. As shown in Example (\ref{ex2}) telugu is a free word order language. Later, \cite{Clark:2016} used reinforcement learning to optimize a neural mention ranking model for coreference resolution.
\begin{description}
\label{ex2}
\setlength{\itemsep}{0pt}
  \setlength{\parskip}{0pt}
  \setlength{\parsep}{0pt}
\item {
Example 2: 

Ram gave Nikhil a book.

S1: rAmu nikhilki pustakam icchADu

(Ram Nikhil book gave)

S2: rAmu pustakam nikhilki icchADu.

(Ram book Shyam gave)

Here the order of the words doesn't affect the meaning of the Telugu sentence.}
\end{description}

\section{Anaphora Resolution in Telugu Language}
 In Telugu, the verbs are formed by adding the grammatical information as suffixes. Along with gender, number and pronoun (GNP), the verb also agrees with tense, aspect, and modality (TAM), which makes the complete structure of the verb as verb root + TAM suffix + GNP suffix. The pronoun should agree with all the components in order to refer to an entity in previous utterances. There are three genders (male, female, nonhuman), three persons (first, second and third) and two numbers (singular and plural) in Telugu. Example (\ref{ex3}) shows the variations produced by changing GNP variables for a common root word 'icchaa'(gave). The subject verb agreement becomes more complex because of honorifics, proximity and formality features attached to the subject in Telugu culture \cite{Subbarao_2000}.

\begin{description}
\label{ex3}
\setlength{\itemsep}{0pt}
  \setlength{\parskip}{0pt}
  \setlength{\parsep}{0pt}
  \item{Example 3: 
  
  For Verb 'gave' when subject is:
\begin{tabular}{lclclclc}
Male&1st&singular:&icchaanu \\
Male&2nd&singular:&icchaavu\\
Male&3rd&singular:&icchaaDu \\
Female&3rd&singular:&icchindi \\
Any &3rd&plural: &iccharu \\

\end{tabular}
}
\end{description}

\subsection{Types of Anaphora}
When two or more entities refer to the same person or thing then it is known as coreference \cite{yale:97,Jurafsky:2000:SLP:555733}. Coreference is of two types exophoric and endophoric. In Exopheric coreference, words or entities refer to something which is outside text or discourse. Whereas in Endophoric coreference, entities refer to words which are present in the text. Endophoric coreference is further divided into two types: Anaphora and Cataphora.

In anaphoric reference, words refer to entities which are earlier mentioned in the discourse, whereas in cataphoric reference words refer to entities which are mentioned later in discourse. Anaphoric references are of different types such as repeated, pronominal, lexical and one anaphora.

\subsection{Types of Pronouns in Telugu}
There is a wide variety of pronouns in Telugu. These pronouns differ in their usage based on gender, number, person or other semantic variables. Listed below are few commonly used types of pronouns in Telugu:

\begin{itemize}
\setlength{\itemsep}{0pt}
  \setlength{\parskip}{0pt}
  \setlength{\parsep}{0pt}
    \item \textbf{Personal Pronouns}: Telugu pronouns that are used as substitutes for known noun phrases. Ex:
    nEnu (I), manamu (we), nIvu (you), vAru (they).
    \item \textbf{Interrogative Pronouns}: Telugu pronouns that indicate questions. Ex: EmI (what), Edi (which), EvaDu (who).
    \item \textbf{Possessive Pronouns}: Telugu pronouns that indicate ownership. Ex: nA (my), atani (his), Amedi (Hers).
    \item \textbf{Adverbial Pronouns}: Telugu adverbs that are formed by combining a pronoun with a preposition. Ex: imducEta  (whereby), anduvaLa (whereby),  imdulO (wherein).
    \item \textbf{Reflexive Pronouns}: These pronouns are used when subject and object are same in a sentence. Ex: tAnu (oneself), tAmu (themself).
    \item \textbf{Demonstrative Pronouns}: Pronouns that point to specific things. Ex: I (This | These), A (That | Those).
    \item \textbf{Reciprocal Pronouns}: Reciprocal pronouns are used to indicate that both the parties got benefited by performing certain action or task. Ex: Okarikokaru (Each other).

\end{itemize}
\section{Methodology}
According to Clark\shortcite{Clark:2015}, the primary motive of a neural mention pair model is to perform a binary classification, predicting whether two vectors are co-referent or not. The vectors should be able to learn the linguistic phenomena that appears in the nominal and pronominal mentions in the dialogues. We call these linguistic devices as features. Since Telugu is verb-final language and verbs are strongly inflected than in English, the noun and verb mentions agree more on gender, number, and person. Therefore, in contrast to the 17 features applied by Clark\shortcite{Clark:2015}, we suggest only 6 features:
\begin{itemize}
\setlength{\itemsep}{0pt}
  \setlength{\parskip}{0pt}
  \setlength{\parsep}{0pt}
    \item Word embeddings (100 Dim)
    \item Gender, Number, Person (10 Dim)
    \item Part-of-Plural (1 Dim)
    \item Speaker-Hearer (2 Dim)
\end{itemize}

This section introduces our framework to build the feature vectors and the deep learning model which associates anaphora with its antecedent. Our methodology can be mainly classified into three stages:
\begin{itemize}
\setlength{\itemsep}{0pt}
  \setlength{\parskip}{0pt}
  \setlength{\parsep}{0pt}
    \item Parsing the dialogues
    \item Feature vector generation
    \item Neural network model
\end{itemize}

\subsection{Parsing the dialogues}
\label{sec:vg}
As Telugu is an agglutinative language to get the mentions from the utterances, we need to use a tokenizer and a sandhi splitter which breaks the complex terms into individual stems or root words. Then use a parts of speech tagger to detect the mentions. Then we need to do morph analysis of each word to extract the Gender, Number and Person features from Telugu dialogues. We used an online shallow parser build by LTRC center at IIIT Hyderabad. This shallow parser takes a text sentence as an input in the form of UTF-8 or WX format and generates an output in the form of Shakti Standard Format (SSF) given by \cite{Bharati:14}. This SSF acts as a common format of data for all the Indian languages. See example 4 for output in SSF format.
\begin{description}
\setlength{\itemsep}{0pt}
  \setlength{\parskip}{0pt}
  \setlength{\parsep}{0pt}
  \item{Example 4:
  
unnADu     	VM    $<$fs af='unDu,v,m,sg,3,,A,A' name="$unnaaDu$"$>$
}

\end{description}

\noindent In the above example, we are able to capture parts of speech of the given word which is 'VM' gender which is ‘m’(Male), number which is ‘sg’(Single) and person which is 3(Third Person). Gender is of three types $any$, $male$, and $female$. Number is of 3 types $zero$, $singular$ and $plural$. Person is of three types $none$, $1st$, $2nd$ and $3rd$. To encode these into the vector we need to hot encode them. So the GNP vector will be a vector of 10 dimensions. In this way, we are extracting three important features of our model i.e., Gender, Number and Person. The shallow parser also helps us the nouns, pronouns and verb phrases in the dialogues, which are potential mentions of real entities.

\subsection{Feature Vector Generation}
For generating the word embeddings for Telugu, we scraped Telugu pages in Wikipedia and Andhrajyothi newspaper. From Andhrajyothi website we scrapped all the telugu articles published between 2015 and 2017. This accounted to a total of around 133148 articles. Using Gensim, a word representation tool, we trained our own word2vec model using the scraped data. After training, we obtained 23,000 unique words (types) in our vocabulary. Each vector is of 100 dimensions. Since the data collected from these sources is vast and a mixture of several domains, the vectors have a rich semantic description. Since the conversations involve plenty of 1st and 2nd person mentions, we suggest an experimental feature called $Speaker-Hearer$. It easily discriminates between the two actors by assigning its value to $-1, 1$ respectively. Plural mention discontinuity is popular in coreference resolution systems, but no work has tackled it. Here we introduce a feature called $Part-of-Plural$ that will allow the model to treat plural definite noun mentions as single mentions. For each mention in the dialogue, we will generate the feature vector by appending all the features making it a 113-dimensional vector.

\subsection{Neural Network Model}
The model we build is a Binary Classification Multilayer Perceptron that classifies the pair as a true or false antecedent and anaphora pair. The input is a feature vector that is created by appending vectors of two mentions making it a 226-dimensional mention pair vector. Given the small dimension of the input, there is no expensive computation involved. So we are using a dense neural network.

\subsubsection{Architecture}
\begin{figure}[htp]
\centering
\includegraphics[width=6cm]{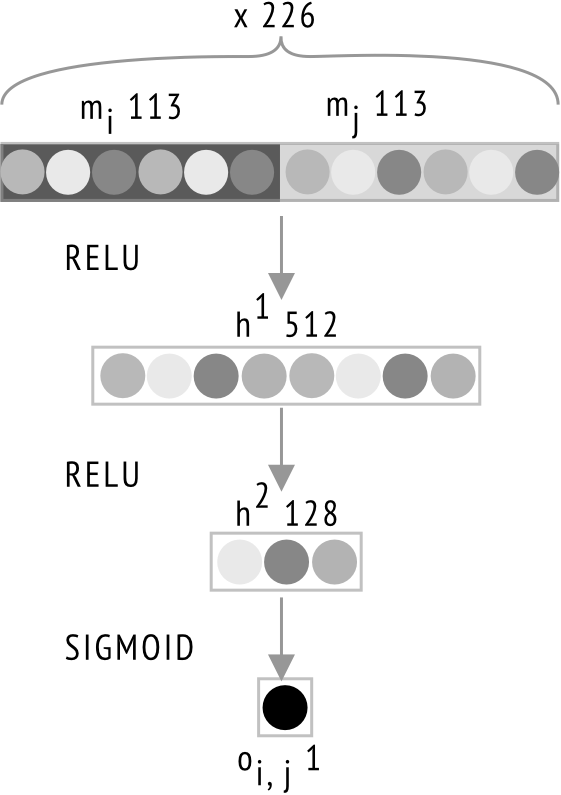}
\caption{Neural Network Model}
\label{fig:model}
\end{figure}

Let $m_i$ be the mention feature vector of the mention $i$ and $p_{(i,j)}$ be the mention pair vector that represents the antecedent-anaphora pair. Now we will send this $p_{(i,j)}$ vector into a fully connected dense neural network with two hidden layers. 
\begin{flalign*}
&Input Layer: x = p_{(i, j)} = [ m_i, m_j ]& \\
&Hidden Layer 1: h^1 = relu(w^1x + b^1)& \\
&Hidden Layer 2: h^2 = relu(w^2h^1 + b^2)& \\
&Output Layer: o_{(i, j)} = sigmoid(w^Th^2)&
\end{flalign*}
The output layer consists of a single value which denotes the probability of the pair to be a true antecedent-anaphora pair. We calculate the loss using a binary cross entropy function.
\begin{flalign*}
	&L(\theta) = -\sum_{i,j\epsilon M,i<j}( y_{(i, j)}\log(o_{(i, j)})+(1-y_{(i, j)})&\\
	&\log(1- o(i, j)))&
\end{flalign*}
Here $M$ are all the annotated mentions in the data set and $y_{(i,j)}$ represents the actual labels of the mention pairs. Here $0$ represents a false pair and $1$ represents a true pair. See figure \ref{fig:model} for the complete model.
\subsubsection{Hyper Parameters}
After each hidden layer, a dropout layer of $0.5$ probability for regularization is added. Regularization helps in over-fitting of the model. Then each epoch of the training phase is optimized using the Adam optimizer \cite{kingma2014adam}. Adam is a momentum based gradient descent optimization technique. We are using a mini-batch of size $128$ pairs in each training epoch. The first hidden layer has $512$ units and the second hidden layer has $128$ units. We use Rectified Linear Unit $(relu)$ activation functions in both the hidden layers and Sigmoid for the last layer. 

\section{Corpus and Annotation}
Telugu is a digital resource-limited language. Most of the research for Telugu was done in sentiment analysis, POS tagging, NER, and text summarization. Publicly available annotated dialogue dataset for Telugu is not available. However, we built a corpus of 157 conversations, consisting of simple to complex dialogues that we hear in our daily life. We collected the corpus in such a way that it consists of all the possible pronoun types and mentions are balanced in gender, number, and person. About 50\% of the conversations are hand engineered, and the remaining 50\% is a translation from English and online scraping. To translate conversations from English to Telugu we are using Google translate API and on top of it a reviewer will evaluate the correctness of the translation, These conversations are then parsed using the shallow parser discussed in section \ref{sec:vg}. The total number of mentions in the corpus is 775. 

After the corpus is ready, the conversations are annotated using a web application we have built specifically for annotating the mentions. The annotator allows you to make a pair of antecedent and anaphora mentions in the conversation. If both the mentions are a single real entity, then they are labeled true, else, they are labeled false. There are 642 true mention pairs and 1818 false mention pairs. The total number of mention pairs in the corpus after oversampling is 3636. Note that the LTRC shallow parser for Telugu is far from human-level performance. So, for enhancing training, the semantic features are corrected and manually tagged with the help of annotator. Each conversation is annotated by two reviewers and in case if there is any conflict, then the conversation is sent to a third reviewer. 

\section{Results}
\begin{figure}[htp]
\centering
\includegraphics[width=6cm]{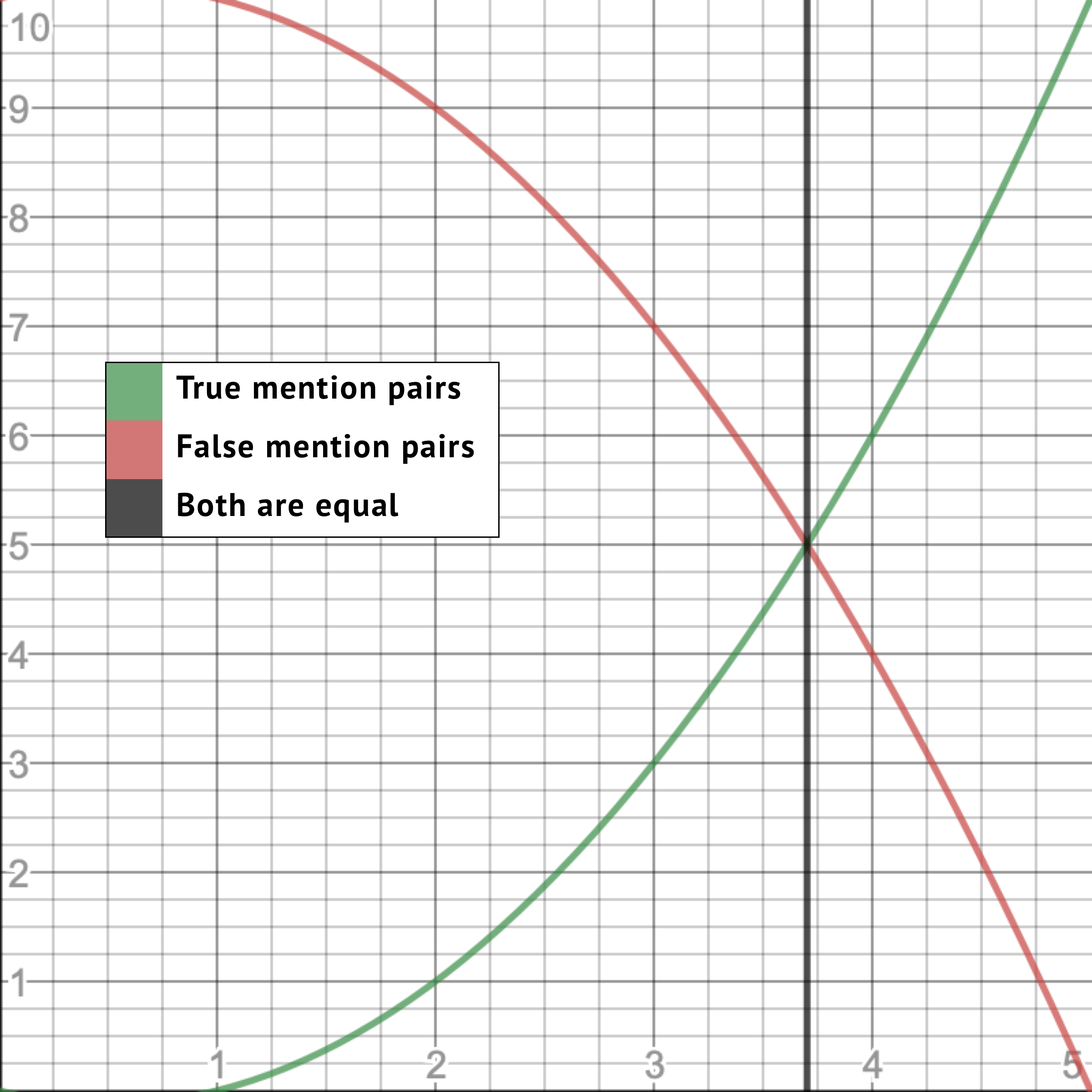}
\caption{True vs False mention pairs}
\label{fig:graph}
\end{figure}

Consider that, in a given context, if there are $n$ mentions, where $n\geq2$, $k$ mentions among them are referring the same entity, where $0\leq k\leq n$. Then there are $k(k-1)/2$ pairs which are true coreference mention pairs and $(n-k)(n+k-1)/2$ pairs which are false coreference mention pairs. After observing the graph constructed based on these two equations for a given $n=5$ and  $0\leq k\leq 5$, there are more possibilities of the false pairs dominating the true pairs. In figure \ref{fig:graph}, we can interpret from the region bounded by the two curves that the true and false mention pairs are unbalanced. This leads to bias while training the model on this corpus.

To fix this we followed sampling strategies. There are two strategies for balancing the data. In undersampling, we will reduce the number of false pair instances randomly. In oversampling, we inflate the number of true pair instances, by generating synthetic samples using a distance-based technique called SMOTE \cite{Chawla_2002}. For testing, a separate set of dialogues are used. See the comparison of the model for both the strategies in table \ref{tab:1}.

To check the performance of the model with features as part of the embedding, we compared the model to the baseline model. A baseline model is a naive model assuming to be the least possible intelligent system. Here we achieved the baseline model by training the neural network only on the 100-dimensional word embeddings. To understand the significance of every feature, we trained the model considering a feature at a time. See table \ref{tab:2} for the comparison based on features.
\begin{table}
\begin{tabular}{ |c|c|c|c|c| } 
\hline
Sampling & Loss & Precision & Recall & F1\\
\hline
Under & 1.8\% & 50.4 & 42.8 & 43.8\\ 
Over & 0.6\% & 83.3 & 90.0 & \textbf{86.0}\\
\hline
\end{tabular}
\caption{Comparison between the sampling strategies}
\label{tab:1}
\end{table}
\begin{table}
\begin{tabular}{ |c|c|c|c|c| } 
\hline
Features & Loss & Precision & Recall & F1\\
\hline
None & 0.9\% & 67.6 & 79.1 & 71.3 \\
Gender & 0.7\% & 80.09 & 86.8 & \textbf{82.5} \\
Number & 0.7\% & 78.2 & 85.7 & 80.8 \\
Person & 0.7\% & 80.0 & 86.2 & 82.3 \\
PoP & 0.7\% & 76.9 & 85.2 & 79.7\\
\hline
\end{tabular}
\caption{Comparison based on features}
\label{tab:2}
\end{table}

\section{Issues}
\subsection{Reporting Speech}

The word vector representation we chose cannot deal with reporting speech. See example (5).
\begin{description}
\setlength{\itemsep}{0pt}
  \setlength{\parskip}{0pt}
  \setlength{\parsep}{0pt}
\item {
Example 5: 

Speaker: Ram said, ‘I am the king of the world’.

Here the pronouns 'I' refers to Ram. But our feature representation will refer to speaker because it is 1st person.}
\end{description}
\subsection{Parser}
When using the system in real conversations, the parser may not give correct GNP tags. These affects the predictions. Also, the morph analyzer gives unnecessary tokenization which leads to unresolved mentions.  
\subsection{Sandhi}
Sometimes the pronoun will be a part of the compound word, which is difficult to split with any computational sandhi splitter in Telugu. 
\begin{description}
\setlength{\itemsep}{0pt}
  \setlength{\parskip}{0pt}
  \setlength{\parsep}{0pt}
\item {
Example 6: 

Only he came.

atanokkaDocchADu

atanu + okkaDu + vacchADu

he   +  alone   +  came
          
Here `he` is part of the compound word which cannot be split and resolved.
}
\end{description}
\section{Conclusion and Future work}
This model is the best anaphora resolution system for Telugu dialogues. It can be used to build more natural conversational agents in Telugu. Since most of the linguistics of the Dravidian language family are similar, we can extend this work for other south Indian languages. The feature vectors are constructible for any language. Our system has surpassed the recent state of the art in Telugu anaphora resolution \cite{Hemanth:15}, whose accuracy is 61.1\%. With more data and discovering more useful features we can further improve this system. 





\bibliographystyle{acl_natbib}

\end{document}